%% file: paper.tex
\documentclass{article}
\usepackage[preprint]{spconf}
\toappear{Submitted to the {\it IEEE ICIP 2021, September 19-22, 2021, Anchorage, Alaska, USA}}
\usepackage{spconf,amsmath,graphicx}
\usepackage{amsfonts} 
\usepackage {tikz}

\newcommand{\abs}[1]{\left\lvert#1\right\rvert} 
\newcommand{\at}{\makeatletter @\makeatother}

\title{D3DLO: Deep 3D LiDAR Odometry}
\name{Philipp Adis, Nicolas Horst, Mathias Wien \thanks{\hspace*{-1.8em}This work is accepted for publication at the IEEE ICIP 2021.\newline \hspace*{1.8em}\copyright\ IEEE 2021. Personal use of this material is permitted.  Permission from IEEE must be obtained for all other uses, in any current or future media, including reprinting/republishing this material for advertising or promotional purposes, creating new collective works, for resale or redistribution to servers or lists, or reuse of any copyrighted component of this work in other works.}}
\address{Institute of Imaging and Computer Vision, RWTH Aachen University, Germany\\\{firstname\}.\{lastname\}\at lfb.rwth-aachen.de\\}

\begin{document}
\ninept
\pagestyle{empty}
\maketitle
\begin{abstract}
LiDAR odometry (LO) describes the task of finding an alignment of subsequent LiDAR point clouds. This alignment can be used to estimate the motion of the platform where the LiDAR sensor is mounted on.
Currently, on the well-known KITTI Vision Benchmark Suite state-of-the-art algorithms are non-learning approaches.
We propose a network architecture that learns LO by directly processing 3D point clouds.
It is trained on the KITTI dataset in an end-to-end manner without the necessity of pre-defining corresponding pairs of points.
An evaluation on the KITTI Vision Benchmark Suite shows similar performance to a previously published work, DeepCLR \cite{deepclr_2020}, even though our model uses only around 3.56\% of the number of network parameters thereof.
Furthermore, a plane point extraction is applied which leads to a marginal performance decrease while simultaneously reducing the input size by up to 50\%.
\end{abstract}
\section{Introduction}
\label{sec:intro}
In robotics and autonomous driving, it is desirable to have accurate estimations of the moving platform's 3D position and orientation, as this provides important information for safe navigation.
Usually, such moving platforms acquire data for the perception of the environment and use on-board sensors such as LiDAR sensors, Inertial Measurement Units, or cameras for motion estimation.
LiDARs are insensitive to lighting conditions and the collected data contains accurate distance information in the form of point clouds.
In recent years, more advanced and cheaper LiDARs have been developed.
Therefore, LiDARs have gained popularity for such applications.

Point Cloud Registration (PCR) is the task of finding an alignment of point clouds by estimating their relative rigid transformation.
Applying PCR to estimate the relative motion between two consecutive LiDAR point clouds is called LiDAR odometry (LO).
A common algorithm for PCR is Iterative Closest Point (ICP) \cite{besl_method_1992}.
It is a point-based algorithm that uses the coordinates of the input points and estimates a rigid transformation by minimizing the distance between corresponding points.
Feature-based approaches, on the other hand, commonly subsample the input point clouds and estimate the transformation based on extracted local features.
In general, they are less sensitive to the quality of the point clouds and therefore more powerful.
One drawback, however, is the sensitivity of feature-based approaches to dynamic objects in the scene \cite{li_lo-net_2020}.
Furthermore, in recent years deep learning has been successfully applied in several computer vision tasks and achieved state-of-the-art performances.
Especially, architectures applying Convolutional Neural Networks (CNNs) for image-based tasks. Unfortunately, the exploration of CNNs in 3D geometric data processing tasks, such as LO, has not been that successful yet.
Therefore, most recent publications for LO use 2D projections of the point clouds and apply CNNs to extract features followed by a transform regression using Multilayer Perceptrons (MLPs) \cite{li_lo-net_2020, wang_deeppco_2020, cho_deeplo_2019, velas_cnn_2017, nicolai2016deep, skikos946end}.
However, as LiDAR point clouds are sparse, the 2D projection can result in empty pixels, which can be circumvented by interpolation.
This, however, can distort the data interpretation significantly.
Moreover, the projection of a 3D point cloud onto a 2D plane can also lead to information loss as several points might fall into one pixel.
PointNet \cite{pointnet} is a pioneering deep learning framework that directly operates on raw point clouds without using convolutions.
It is a baseline for several recently published architectures for tasks such as point cloud semantic segmentation or scene flow estimation \cite{pointnet_pp, flownet3d}. 

%
Based on the findings of previous work, which we will summarize in section \ref{sec:rel_work}, the main part of this work is dedicated to the introduction of our proposed model in section \ref{sec:method}.
It is similar to \cite{deepclr_2020} and performs comparably while using only about 3.56\% the number of trainable parameters thereof.
%
In sections \ref{sec:training} and \ref{sec:evaluation} the training and evaluation of the model are introduced.
We extract plane points of the input point clouds which reduces the data by up to 50\% in size to analyze how this affects the network's performance.
Finally, in sections \ref{sec:results} and \ref{sec:conclusion} we will present the evaluation results and come to a conclusion about our work.

\section{Related Work}
\label{sec:rel_work}
%
Given two point clouds and an initial guess for the target transform, the ICP algorithm \cite{besl_method_1992} iteratively refines this transform by defining corresponding pairs of points and minimizing the overall distance between them.
Further development of the algorithm was achieved e.g. by applying a modified distance metric as done in \cite{segal2009generalized}.
Unfortunately, in pure LO no initial guess of the target transformation is given.
This can lead to large inaccuracies when applying ICP at high velocities of the moving platform due to the large relative displacement of subsequent point clouds \cite{deepclr_2020}.
In recent years LOAM \cite{zhang_loam_2014} has been considered as a state-of-the-art approach for LO and is also ranked as the best LO-method on the KITTI Vision Benchmark Suite. Two algorithms operate in parallel at frequencies of 10 \nolinebreak Hz and 1 Hz. The first estimates the current velocity and corrects distortions of the point clouds that occur due to ego-motion. The second algorithm uses the undistorted point clouds for registration and mapping.
%

Promising results were achieved applying deep learning to tasks such as visual odometry (VO) \cite{wang2018end, Yang_2020_CVPR}, image-based localization or pose estimation \cite{Kendall_2015_ICCV, Kendall_2017_CVPR}, and point cloud classification and segmentation \cite{pointnet, pointnet_pp, NEURIPS2018_f5f8590c}.
However, using deep learning to process 3D geometric data for tasks such as LO has not been that successful as of now.
The majority of deep learning methods for LO project the LiDAR data onto a spherical or cylindrical plane around the sensor and use the LiDAR intensity, depth, XYZ coordinates or normal vectors as channel values for each pixel \cite{li_lo-net_2020, wang_deeppco_2020, cho_deeplo_2019, velas_cnn_2017, nicolai2016deep, skikos946end}.
CNNs are commonly used to extract features of subsequent frames followed by a single- \cite{li_lo-net_2020, velas_cnn_2017, nicolai2016deep} or dual-branch \cite{wang_deeppco_2020} MLP for a joint or separate estimation of translational and rotational target values.
\cite{cho_deeplo_2019} uses two individual sub-networks to extract translational and rotational information based on residual blocks with fully convolutional networks.
The extracted information is fed into a PoseNet \cite{Kendall_2015_ICCV} to regress the translational and rotational outputs.

DeepCLR \cite{deepclr_2020} applies a multi-scale set abstraction layer to subsample individual raw 3D point clouds and simultaneously extract features similar to \cite{pointnet_pp}.
Subsequently, a flow embedding layer is used to learn flow embeddings based on the extracted features of two consecutive subsampled point clouds as suggested in \cite{flownet3d} for scene flow estimation.
These learned embeddings are then fed into an MLP to regress the target transform.
The architecture of our model is based on this approach but with 61,290 parameters it uses only around 3.56\% of the number of the trainable network parameters of DeepCLR as shown in Table \ref{tab:model} in section \ref{subsec:architecture}.
\section{Method}
\label{sec:method}
Our model uses the same types of layers as DeepCLR \cite{deepclr_2020}.
The main difference lies in the number and order of applied layers and their parameterization.
\subsection{Problem Definition}
Using two point clouds $P$ and $Q$ corresponding to time steps $t$ and $t+1$ as input, we want to find the homogeneous transformation $T_{t,t+1} \in \mathbb{R}^{4\times4}$ between them.
$T_{t,t+1}$ aligns the poses $T_{0,t}$ and $T_{0,t+1}$ of $P$ and $Q$ according to eq.~(\ref{eq:1}), with the pose at time step 0 defining the global coordinate system.
$F$ and $\theta$ denote the function modeled by the network and the set of learned network parameters and $\hat{T}_{t, t+1} = F(P,Q;\theta)$ is the network output.
\begin{equation}\label{eq:1}
    T_{0,t+1} = T_{0,t} \cdot T_{t,t+1}
\end{equation}
In the following, $T_{t,t+1}$ and $\hat{T}_{t, t+1}$ will simply be referred to as the target transform $T$ between two subsequent point clouds and the corresponding prediction of the model $\hat{T}$.
\subsection{Output Representation and Objective Function}
The target transformation $T$ consists of a translational part $t \in \mathbb{R}^{3}$ and a rotation matrix $R \in \mathbb{R}^{3\times3}$.
The rotational part is commonly represented using Euler angles \cite{wang_deeppco_2020,skikos946end} or quaternions \cite{li_lo-net_2020, cho_deeplo_2019} rather than a rotational matrix which results in fewer output parameters and circumvents the orthogonality constraint of rotation matrices \cite{Kendall_2017_CVPR}.
We represent the predicted transformation $\hat{T}$ as a translation $\hat{t} \in \mathbb{R}^{3}$ in meters and a rotation $\hat{r} \in \mathbb{R}^{3}$ in Euler angles in degrees as this choice leads to similar value ranges of the target translations and rotations.
The application of quaternions was also investigated but did not lead to an improved performance.

Due to the choice of the output representation we apply the Mean Absolute Error (MAE) as objective function without an individual weighting of translational and rotational outputs as shown in eq.~(\ref{eq:unweighted_loss}) with $y=\left[t \; r\right]^T, \hat{y}=\left[\hat{t} \; \hat{r} \right]^T \in \mathbb{R}^{6}$.
Homoscedastic uncertainty weighting \cite{Kendall_2017_CVPR} was investigated but did not show any improvement.
\begin{equation}\label{eq:unweighted_loss}
    L_{\text{unweighted}} = \frac{1}{6} \sum_{i=0}^{5}\abs{y_i-\hat{y}_i}
\end{equation}
\subsection{Network Architecture}\label{subsec:architecture}
\begin{figure}
    \centering
    \input{images/architecture}
    \caption{Illustration of developed model.}
    \label{fig:network}
\end{figure}
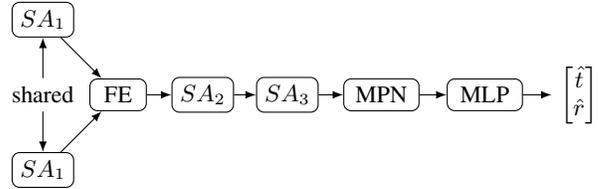

In a first step, a shared set abstraction (SA) layer \cite{pointnet_pp} simultaneously subsamples two consecutive point clouds $P\in \mathbb{R}^{n_1 \times (3+c)}, Q \in \mathbb{R}^{n_2 \times (3+c)}$ and learns to aggregate local feature vectors for the remaining points.
We use Cartesian coordinates and the LiDAR intensity as input which results in $c=1$.
A flow embedding (FE) layer \cite{flownet3d} learns features containing information of the flow of each point $\{p_i\}_{i=1}^{n_1} \in P$.
Afterwards, two more SA layers are applied to further subsample the remaining points of the first point cloud and their assigned flow embeddings.
Before regressing the final transformation with an MLP a mini-PointNet (MPN) \cite{deepclr_2020} is applied to combine the information of all remaining points into one global feature vector.
This step is the key to keeping the overall number of parameters small as it circumvents the concatenation of the remaining points, and consequently allows to use a shallow MLP to regress the target transformation. An overview is given in Fig. \ref{fig:network}.
\subsubsection{Mini-PointNet}
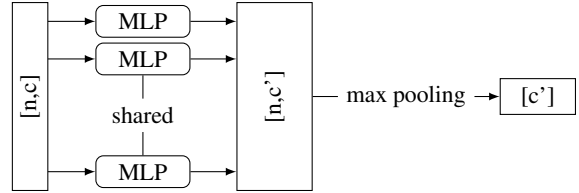
\begin{figure}
    \centering
    \input{images/mpn}
    \caption{Illustration of the mini-PointNet \cite{deepclr_2020}. It transforms n input feature vectors of dimension c into dimension c' using a shared MLP. Subsequent max pooling along the point dimension n results in one global feature vector of dimension c'.}
    \label{fig:mpn}
\end{figure}

In this work, the mini-PointNet \cite{deepclr_2020} is used to learn the information aggregation of an arbitrary number of local feature vectors into one global feature vector.
A non-linear function $h_{\text{mpn}}: \mathbb{R}^{c} \rightarrow \mathbb{R}^{c'}$ transforms the local feature vectors from dimension $c$ into dimension $c'$.
Subsequent element-wise max pooling results in one output vector of dimension $c'$. The function $h_{\text{mpn}}$ is modeled with an MLP.
Fig. \ref{fig:mpn} illustrates this process.
\subsubsection{Set Abstraction Layer}
\begin{figure}
    \centering
    \def\tkzscl{0.025}
    \input{images/sa_layer}
    \caption{Illustration of the SA layer. Using the Farthest Point Sampling (FPS) algorithm $n_{\text{fps}}$ (here $n_{\text{fps}}=2$) centroids (gray) are defined. Using the XYZ coordinates ($\mathbb{R}^{3}$) and features ($\mathbb{R}^{c}$) of the centroids and their neighboring points new features ($\mathbb{R}^{c'}$) are learned by applying a shared mini-PointNet. As output only the centroids' coordinates and the learned features are kept.}
    \label{fig:sa_layer}
\end{figure}
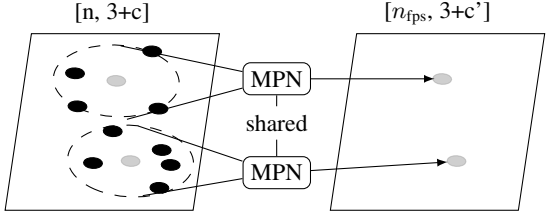

\begin{table}[t]
	\centering
	\begin{tabular}{ |p{1.05cm}||p{1cm}|p{1cm}|p{0.5cm}|p{1.8cm}|  }
		\hline
		Layer       & r [m] & $ n_{\text{fps}}$& $ n_n $& MLP \\
		\hline  
		$SA_1$			& 1.0   & 1024      & 8      & [4, 8, 16, 32]      \\
		FE			& -     &  -        & 16     & [32, 64]  \\
		$SA_2$			& 4.0   & 256       & 32     & [64, 64]  \\
		$SA_3$			& 8.0   & 64        & 8      & [64, 64]  \\
		MPN		    & - 	& -		    & -      & [64, 256] \\
		MLP 	    & -		& -	        & -      & [64, 6]\\
		\hline
	\end{tabular}
	\caption{Model definition with 61{,}290 parameters.}
	\label{tab:model}
\end{table}
The set abstraction layer \cite{pointnet_pp} takes a point cloud $P=\{p_i\}_{i=1}^{n} \in \mathbb{R}^{n\times(3+c)}$ with $p_i=\{x_i,f_i\}$ consisting of XYZ coordinates $x_i \in \mathbb{R}^{3}$ and features $f_i \in \mathbb{R}^{c}$ as input.
Applying the Farthest Point Sampling (FPS) algorithm a pre-defined number of $n_{\text{fps}}$ centroids are defined with $n_\text{fps} \leq n$.
The output of the layer comprises a subsampled point cloud $P'=\{p_j'\}_{j=1}^{n_{\text{fps}}} \in \mathbb{R}^{n_{\text{fps}} \times (3+c')}$ with its XYZ coordinates $x_j' \in \mathbb{R}^{3}$ and features $f_j' \in \mathbb{R}^{c'}$.
In a grouping process for each centroid, neighboring points within a pre-defined radius $r$ are gathered.
Using a non-linear function $h_{\text{sa}}: \mathbb{R}^{3+c} \rightarrow \mathbb{R}^{c'}$ which is modeled with an MLP, and subsequent element-wise max pooling a new feature is learned for each centroid $p_j'$ according to eq.~(\ref{eq:set_ab_layer}):
\begin{equation}\label{eq:set_ab_layer}
f_j' = \max_{\{i: \left\Vert x_i-x_j' \right\Vert_2 \leq r\}} \{h_{\text{sa}}(f_i, x_i-x_j')\}
\end{equation}
%
By only keeping the centroids' XYZ coordinates and the aggregated feature and discarding all the neighboring points, a subsampling and simultaneous feature extraction is achieved.
For an efficient implementation, a maximum number of neighbors $n_n$ is required to limit the number of neighboring points during the grouping process.
Furthermore, we handle the edge case of no neighboring points being present within radius $r$ by choosing the centroid itself as a neighbor. Fig. \ref{fig:sa_layer} shows the working principle of the defined SA layer.
As shown in \cite{deepclr_2020, pointnet_pp} it is possible to apply several SA layers with individual radii in parallel.
The features resulting from the individual radii are concatenated for each centroid.
Such an SA layer is called a multi-scale grouping (msg) SA layer and is ought to extract features of objects of different scales.
The application of an msg-SA layer as input layer was investigated in this work but did not come with any performance improvement.
\subsubsection{Flow Embedding Layer}
\begin{figure}
    \centering
    \def\tkzscl{0.025}
    \input{images/fe_layer}
    \caption{Illustration of the FE layer. To each point of the first input point cloud $P \in \mathbb{R}^{n_1 \times (3+c)}$ (dark blue) and the corresponding neighboring points within the second input point cloud $Q \in \mathbb{R}^{n_2 \times (3+c)}$ (black) an MPN is applied. This way a flow embedding for all points in $P$ (here shown for two points colored in gray) is learned which describes the motion between $P$ and $Q$. The output $P' \in \mathbb{R}^{n_1 \times (3+c')}$ (light blue) consists of the XYZ coordinates ($\mathbb{R}^3$) of each point in $P$ and the learned flow embeddings ($\mathbb{R}^{c'}$) as features.}
    \label{fig:fe_layer}
\end{figure}
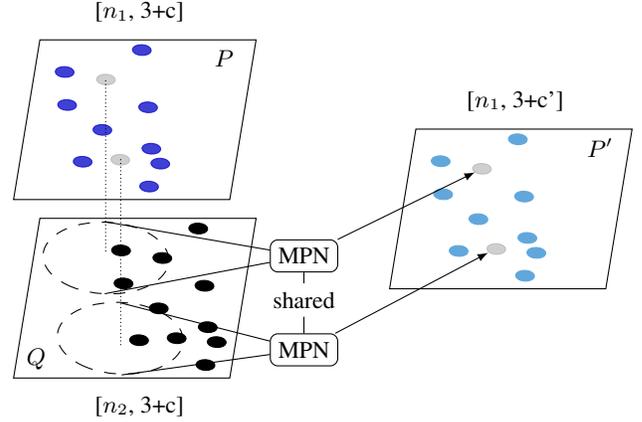

The flow embedding layer as introduced in \cite{flownet3d} functions similar to the SA layer.
Instead of a single one, it takes two point clouds $P=\{p_i\}_{i=1}^{n_1} \in \mathbb{R}^{n_1\times(3+c)}$ and $Q=\{q_j\}_{j=1}^{n_2} \in \mathbb{R}^{n_2\times(3+c)}$ as input.
The points $p_i=\{x_i,f_i\}$ and $q_j=\{y_j,g_j\}$ come with their XYZ coordinates $x_i,y_j \in \mathbb{R}^{3}$ and features $f_i,g_j \in \mathbb{R}^{c}$.
In a grouping process for each point $p_i \in P$, neighboring points $q_j \in Q$ within a pre-defined radius $r$ are gathered.
Similar to the SA layer, a non-linear function $h_{\text{fe}}: \mathbb{R}^{3+2c} \rightarrow \mathbb{R}^{c'}$, modeled with an MLP, and element-wise max pooling are applied to learn a flow embedding $e_i \in \mathbb{R}^{c'}$ for each point $p_i$ according to eq.~(\ref{eq:flow_embed}):
\begin{equation}\label{eq:flow_embed}
e_i = \max_{\{j: \left\Vert y_j-x_i \right\Vert_2 \leq r\}} \{h_{\text{fe}}(f_i, g_j, y_j-x_i)\}
\end{equation}
The output $P' \in \mathbb{R}^{n_1 \times (3+c')}$ of the FE layer is defined as the XYZ coordinates $x_i$ and the learned embeddings $e_i$ of each point $p_i$. Fig. \ref{fig:fe_layer} shows the working principle of such an FE layer.
Due to the edge case where no neighboring points $q_j$ are located within radius $r$ of point $p_i$ we apply a $k$-nearest neighbors search instead of grouping with radius $r$ where $k$ is set to a pre-defined number $n_n$ of neighboring points. 

An overview of the model and the choice of fixed parameters are shown in Table \ref{tab:model}.
One key difference to \cite{deepclr_2020} is the choice of more stacked layers applying rather shallow MLPs which leads to an overall architecture with around 3.56\% the number of network parameters of \cite{deepclr_2020}.
The radii and values of $n_n$ for the SA and FE layers are chosen based on the point distribution of the point clouds in the KITTI dataset.
\subsection{Regularization}\label{subsec:reg}
As suggested in \cite{flownet3d_pp} for the task of scene flow estimation we investigate regularization by adding the cosine distance $L_{\textrm{cos\_dist}}$ between the ground truth $y = [t, r]^T$ and prediction $\hat{y} = [\hat{t}, \hat{r}]^T$, according to eq.~(\ref{eq:cos_dist}), as a regularization term to the overall loss function.
\begin{equation}\label{eq:cos_dist}
L_{\textrm{cos\_dist}} = 1 - \frac{y \cdot \hat{y}}{\left\Vert y \right\Vert_{2}  \cdot \left\Vert \hat{y} \right\Vert_{2}}
\end{equation}
This way predictions with a large angle deviation from the ground truth are penalized more than predictions with a smaller angle, even if they deviate the same regarding the euclidean distance.
The regularization with $L_{\textrm{cos\_dist}}$ is tested in a weighted as well as an unweighted manner.
Furthermore, we also analyze the effect of applying the cosine distance to the translational parts of the ground truth $t$ and prediction $\hat{t}$ only.
\section{Training}\label{sec:training}
In the KITTI dataset, about 40\% - 50\% of the points belong to the plane defined by the street.
To analyze the importance of these plane points the training is conducted with raw 3D point clouds and pre-processed point clouds without plane points.
To extract the plane points a RANSAC-based plane estimation of the open-source library Open3D \cite{Zhou2018} is used to define and extract plane points.
Data augmentation is applied by randomly swapping the order of the input point clouds and correspondingly inverting the ground truth transform.
We use ReLU activation and batch normalization is applied in all layers but the final one.
The training is done for 500 epochs using the Adam optimizer with default values \cite{kingma2014adam} and learning rate decays are applied at epochs 300 and 400.
We use sequences 0, 1, 2, 8, 9 for training and 3, 4, 5, 6, 10 for testing during the optimization as done in \cite{skikos946end}.
Sequence 7 is used for validation only.
Furthermore, different values for the network parameter $n_n$ of the first SA layer are investigated to analyze the influence on the network's performance.
%
\section{Evaluation}\label{sec:evaluation}
In odometry, a model's performance on sequences or sub-sequences is of interest rather than its frame-to-frame performance.
In contrast to the training process where the model is optimized frame-to-frame, we evaluate the translation and rotation individually on sub-sequences of certain lengths.
Let $S=\{(i,j)\}$ be a set of tuples of start and end indices $(i,j)$ of all sub-sequences (of sequences 0-10) of lengths 100 m, 200 m, ..., 800 m.
The error transform $T_{\text{error},i,j} = T_{i,j}^{-1} \cdot \hat{T}_{i,j}$ describes the relative transform between the predicted and ground truth end-points of sub-sequence $(i,j)$ and consists of the translation $t_{\text{error},i,j} \in \mathbb{R}^{3}$ and the rotation matrix $R_{\text{error},i,j} \in \mathbb{R}^{3\times3}$ \cite{geiger_are_2012}.
Furthermore, we define the absolute rotation angle $\abs{\theta_{i,j}}$ between the predicted and ground truth end-points according to eq.~(\ref{eq:abs_rot_angle}).
The translational error in percent and rotational error in deg/m are defined according to equations (\ref{eq:trans_error}) and (\ref{eq:rot_error}) with dist$(i,j)$ being the ground truth euclidean distance of sub-sequence $(i,j)$.
\begin{equation}\label{eq:abs_rot_angle}
\abs{\theta_{i,j}} = \textrm{arccos} \left( \frac{\textrm{Tr}(R_{\text{error},i,j})-1}{2} \right)
\end{equation}
\begin{equation}\label{eq:trans_error}
	E_\text{t} = \frac{1}{|S|} \sum_{(i,j) \in S} \frac{\left\Vert t_{\textrm{error},i,j} \right\Vert_2}{\textrm{dist}(i,j)}
\end{equation}
\begin{equation}\label{eq:rot_error}
E_\text{r} = \frac{1}{|S|} \sum_{(i,j) \in S} \frac{|\theta_{i,j}|}{\textrm{dist}(i,j)}
\end{equation}
%
\section{Results}\label{sec:results}
The first SA layer of our model was trained with values for $n_n \in \{8, 16, 32, 64, 128\}$ as shown in Table \ref{tab:kitti_eval_model_all}.
The evaluation was done using the metrics defined in equations (\ref{eq:trans_error}) and (\ref{eq:rot_error}) on sub-sequences of all available sequences (0-10). 
\begin{table}[t]
	\centering
	\begin{tabular}{ |p{0.3cm}||p{1cm}|p{1cm}|p{1cm}|p{1cm}|p{1cm}||p{0.25cm}|}
		\hline
        $ n_n $ 		    & 8  	            & 16	  & 32 		        & 64 	  &  128    &  pp\\
        \hline
		\hline
		$E_\text{t}$ 	        & \textbf{2.19} 	& 2.5  	  & 2.53  	        & 3.63 	  & 3.04     & y\\  	 
		\cline{1-6}
		$E_\text{r}$             & \textbf{0.00855}  & 0.01015 & 0.00937         & 0.01483 & 0.01172  &  \\
		\hline
		\hline
		$E_\text{t}$ 	        & 2.29 	            & 2.76    & \textbf{2.28}  	& 2.52 	  & 2.82     &  n  \\  	 
		\cline{1-6}
		$E_\text{r}$             & \textbf{0.00946}  & 0.01115 & 0.00969         & 0.01057 & 0.01192  &        \\
		\hline

	\end{tabular}
	\caption{Evaluation on all sequences of the model trained with different values of $n_n$ in the first SA layer and point clouds with and without plane points (pp) as input.}
	\label{tab:kitti_eval_model_all}
\end{table}
The evaluation results show that the model achieves similar performance when trained with points clouds with or without plane points.
This is remarkable as the reduced point clouds consist of only 50\% - 60\% of the number of points of the raw point clouds. 
The evaluation results when using only the test and validation sequences for the best three models are shown in Table \ref{tab:kitti_eval_model_test_eval}.
\begin{table}[t]
	\centering
	\begin{tabular}{ |p{0.8cm}||p{1.2cm}|p{1.5cm}|p{1.6cm}||p{1.2cm}| }
		\hline
        $ n_n, pp $ 	& 8, w. pp  	& 8, w/o. pp  & 32, w/o. pp 	& split\\
        \hline
		\hline
		$E_\text{t}$ 	    & \textbf{3.49} 	& 3.98  	 & 4.32  	    & test  \\  	 
		\cline{1-4}
		$E_\text{r}$       & \textbf{0.01393}  & 0.01571    & 0.01642      &  \\
		\hline
		\hline
		$E_\text{t}$ 	    & \textbf{2.5} 	    & 3.89  	 & 4.07  	    & validation  \\  	 
		\cline{1-4}
		$E_\text{r}$         & \textbf{0.01646}  & 0.02563    & 0.03089      &  \\
		\hline
	\end{tabular}
	\caption{Evaluation of the best models when using point clouds with or without plane points (pp) on test and validation sequences.}
	\label{tab:kitti_eval_model_test_eval}
\end{table}
The evaluation results in Table \ref{tab:kitti_eval_model_test_eval} are slightly worse than the results in Table \ref{tab:kitti_eval_model_all}.
Intuitively, this might suggest overfitting and the results on the whole dataset in Table \ref{tab:kitti_eval_model_all} might be lifted up by the model's performance on the training data.
Actually, the contrary could be observed.
The loss during the optimization process on the test data is slightly smaller than on the training data and an even smaller loss results on the validation data after the training.
Larger evaluation errors on the test and validation splits might result from the fact that the models are optimized using the unweighted frame-to-frame metric and not the sequential metrics from equations (\ref{eq:trans_error}) and (\ref{eq:rot_error}).

Regarding the regularization using the cosine distance $L_{\textrm{cos\_dist}}$ (eq.~(\ref{eq:cos_dist})) no performance improvement was achieved which holds for all variations introduced in section \ref{subsec:reg}.
This is in contrast to the findings in \cite{flownet3d_pp} where the cosine distance is successfully applied for the task of scene flow estimation.
In this case, however, a translational vector for each point of the first point cloud instead of a rigid transformation is estimated which leads to different optimization problems. 

The best model with $n_n=8$ and point clouds with plane points as input was submitted to the KITTI Vision Benchmark Suite to be evaluated on the eleven sequences for which no ground truth is publicly available.
The evaluation resulted in $E_t = 5.4\%$ and $E_r = 0.0154$ deg/m and is currently ranked 121/134.
This is remarkable as our model consists of only 3.56\% of the number of parameters of DeepCLR \cite{deepclr_2020} which is ranked 118/134 with $E_t = 4.19\%$ and $E_r = 0.0087$ deg/m (numbers as of the date of paper submission).
%
\section{Conclusion}\label{sec:conclusion}
We presented an end-to-end network architecture for point cloud registration using raw point clouds.
Experiments were done to find suitable fixed network parameters and the best model was submitted to the KITTI Vision Benchmark Suite for evaluation.
The overall architecture is similar to DeepCLR \cite{deepclr_2020} but uses only 3.56\% of the number of network parameters which manifests itself in a slight performance decrease compared to DeepCLR.
Furthermore, we showed that pre-processing the point clouds by extracting plane points and thereby reducing the point clouds in size by up to 40\% - 50\% only leads to a small performance decrease.
This suggests that the necessary information to perform LiDAR odometry might not be contained within the plane points.
Future work could include more investigations on the pre-processing of the point clouds.
Furthermore, considering the sequential information contained in the data by applying Long Short-Term Memory Units (LSTMs) might lead to a performance improvement.

%
%
%
%


\bibliographystyle{IEEEbib}
\bibliography{strings,refs} 

\end{document}

%% file: images/architecture.tex
\tikzstyle{block_small}=[draw, rounded corners=0.1cm, minimum width=0.75cm]
\tikzstyle{block_large}=[draw, rounded corners=0.1cm, minimum width=1cm]
\tikzstyle{arrow} = [->, color=black, >=latex]

\begin{tikzpicture}

    \node[block_small](SA11) at (-1, 1)  {$SA_1$};
    \node [text centered] at (-1, 0) (shared) {shared};
    \node[block_small](SA12) at (-1, -1)  {$SA_1$};
    \node[block_small](FE) at (0, 0)  {FE};
    \node[block_small](SA2) at (1.125, 0)  {$SA_2$};
    \node[block_small](SA3) at (2.25, 0)  {$SA_3$};
    \node[block_large](MPN) at (3.5, 0)  {MPN};
    \node[block_large](MLP) at (4.875, 0)  {MLP};
    \node [text centered] at (6.125, 0) (T) {$\begin{bmatrix}
                                                   \hat{t} \\
                                                   \hat{r}
                                            \end{bmatrix}$};                                    
    
    \draw[arrow] (shared) to (SA11);
    \draw[arrow] (SA11) to (FE);
    \draw[arrow] (SA12) to (FE);
    \draw[arrow] (shared) to (SA12);
    \draw[arrow] (FE) to (SA2);
    \draw[arrow] (SA2) to (SA3);
    \draw[arrow] (SA3) to (MPN);
    \draw[arrow] (MPN) to (MLP);
    \draw[arrow] (MLP) to (T);
    
\end{tikzpicture}


%% file: images/mpn.tex
\tikzstyle{block_large}=[draw, rounded corners=0.1cm, minimum width=1.25cm]
\tikzstyle{anchor_point}=[text centered]
\tikzstyle{arrow} = [->, color=black, >=latex]
\tikzstyle{line} = [-, color=black, >=latex]

\begin{tikzpicture}

    \node[block_large](mlp1) at (0, 0)  {MLP};
    \node[block_large](mlp2) at (0, -0.5)  {MLP};
    \node[block_large](mlpn) at (0, -2)  {MLP};
    \node [text centered] at (0, -1.25) (shared) {shared};
    \node[draw, rotate=90, minimum width=2.5cm] at (-1.5,-1) (inputs) {[n,c]};
    \node[draw, rotate=90, minimum width=2.5cm, minimum height=1cm] at (1.75,-1) (trans_feat) {[n,c']};
    \node[draw, minimum width=1cm] at (5.25,-1) (feat_out) {[c']};
    \node [text centered] at (3.5, -1) (max_pool) {max pooling};
    \node[anchor_point] (p11) at (-1.375,0) {};
    \node[anchor_point] (p21) at (-1.375,-0.5) {};
    \node[anchor_point] (pn1) at (-1.375,-2) {};
    \node[anchor_point] (p12) at (1.355,0) {};
    \node[anchor_point] (p22) at (1.355,-0.5) {};
    \node[anchor_point] (pn2) at (1.355,-2) {};
    \draw[line] (shared) to (mlp2);
    \draw[line] (shared) to (mlpn);
    \draw[line] (trans_feat) to (max_pool);
    \draw[arrow] (p11) to (mlp1);
    \draw[arrow] (p21) to (mlp2);
    \draw[arrow] (pn1) to (mlpn);
    \draw[arrow] (mlp1) to (p12);
    \draw[arrow] (mlp2) to (p22);
    \draw[arrow] (mlpn) to (pn2);
    \draw[arrow] (max_pool) to (feat_out);
\end{tikzpicture}

%% file: images/sa_layer.tex
\tikzstyle{block_large}=[draw, rounded corners=0.1cm, minimum width=0.5cm]
\tikzstyle{arrow} = [->, color=black, >=latex]
\tikzstyle{line} = [-, color=black, >=latex]
\tikzstyle{anchor_point}=[text centered]
\tikzstyle{black_point}=[draw, circle, fill=black, inner sep=0pt, minimum size=0.2cm]
\tikzstyle{grey_point}=[draw, circle, fill=gray!50, inner sep=0pt, minimum size=0.2cm]

\begin{tikzpicture}[scale=\tkzscl]

\draw  [color={rgb, 255:red, 0; green, 0; blue, 0 }  ,draw opacity=1 ][fill={rgb, 255:red, 0; green, 0; blue, 0 }  ,fill opacity=1 ] (82,49.79) .. controls (84.18,48.74) and (87.32,48.86) .. (89.01,50.04) .. controls (90.71,51.22) and (90.32,53.02) .. (88.15,54.06) .. controls (85.98,55.11) and (82.84,55) .. (81.14,53.82) .. controls (79.45,52.64) and (79.83,50.83) .. (82,49.79) -- cycle ;
\draw  [color={rgb, 255:red, 0; green, 0; blue, 0 }  ,draw opacity=1 ][fill={rgb, 255:red, 0; green, 0; blue, 0 }  ,fill opacity=1 ] (46.68,62.99) .. controls (48.86,61.95) and (52,62.06) .. (53.69,63.24) .. controls (55.39,64.42) and (55,66.22) .. (52.83,67.27) .. controls (50.66,68.31) and (47.52,68.2) .. (45.82,67.02) .. controls (44.13,65.84) and (44.51,64.04) .. (46.68,62.99) -- cycle ;
\draw  [color={rgb, 255:red, 0; green, 0; blue, 0 }  ,draw opacity=1 ][fill={rgb, 255:red, 0; green, 0; blue, 0 }  ,fill opacity=1 ] (83.22,69.92) .. controls (85.39,68.88) and (88.53,68.99) .. (90.23,70.17) .. controls (91.92,71.35) and (91.54,73.15) .. (89.37,74.2) .. controls (87.19,75.24) and (84.06,75.13) .. (82.36,73.95) .. controls (80.66,72.77) and (81.05,70.97) .. (83.22,69.92) -- cycle ;
\draw  [color={rgb, 255:red, 155; green, 155; blue, 155 }  ,draw opacity=0.48 ][fill={rgb, 255:red, 155; green, 155; blue, 155 }  ,fill opacity=0.48 ] (66.69,64.05) .. controls (68.87,63.01) and (72,63.12) .. (73.7,64.3) .. controls (75.4,65.48) and (75.01,67.28) .. (72.84,68.33) .. controls (70.67,69.37) and (67.53,69.26) .. (65.83,68.08) .. controls (64.13,66.9) and (64.52,65.1) .. (66.69,64.05) -- cycle ;
\draw  [color={rgb, 255:red, 0; green, 0; blue, 0 }  ,draw opacity=1 ][fill={rgb, 255:red, 0; green, 0; blue, 0 }  ,fill opacity=1 ] (38.42,93.2) .. controls (40.59,92.15) and (43.73,92.26) .. (45.43,93.44) .. controls (47.12,94.62) and (46.74,96.43) .. (44.57,97.47) .. controls (42.39,98.51) and (39.25,98.4) .. (37.56,97.22) .. controls (35.86,96.04) and (36.24,94.24) .. (38.42,93.2) -- cycle ;
\draw  [color={rgb, 255:red, 0; green, 0; blue, 0 }  ,draw opacity=1 ][fill={rgb, 255:red, 0; green, 0; blue, 0 }  ,fill opacity=1 ] (87.98,61.98) .. controls (90.15,60.94) and (93.29,61.05) .. (94.99,62.23) .. controls (96.69,63.41) and (96.3,65.22) .. (94.13,66.26) .. controls (91.96,67.3) and (88.82,67.19) .. (87.12,66.01) .. controls (85.42,64.83) and (85.81,63.03) .. (87.98,61.98) -- cycle ;
\draw  [color={rgb, 255:red, 0; green, 0; blue, 0 }  ,draw opacity=1 ][fill={rgb, 255:red, 0; green, 0; blue, 0 }  ,fill opacity=1 ] (57.17,79.93) .. controls (59.34,78.89) and (62.48,79) .. (64.18,80.18) .. controls (65.87,81.36) and (65.49,83.16) .. (63.31,84.21) .. controls (61.14,85.25) and (58,85.14) .. (56.31,83.96) .. controls (54.61,82.78) and (54.99,80.97) .. (57.17,79.93) -- cycle ;
\draw  [color={rgb, 255:red, 0; green, 0; blue, 0 }  ,draw opacity=1 ][fill={rgb, 255:red, 0; green, 0; blue, 0 }  ,fill opacity=1 ] (81.48,91.97) .. controls (83.65,90.92) and (86.79,91.03) .. (88.49,92.21) .. controls (90.19,93.39) and (89.8,95.2) .. (87.63,96.24) .. controls (85.45,97.29) and (82.32,97.17) .. (80.62,95.99) .. controls (78.92,94.81) and (79.31,93.01) .. (81.48,91.97) -- cycle ;
\draw  [color={rgb, 255:red, 0; green, 0; blue, 0 }  ,draw opacity=1 ][fill={rgb, 255:red, 0; green, 0; blue, 0 }  ,fill opacity=1 ] (78.15,122.44) .. controls (80.33,121.4) and (83.46,121.51) .. (85.16,122.69) .. controls (86.86,123.87) and (86.47,125.68) .. (84.3,126.72) .. controls (82.13,127.76) and (78.99,127.65) .. (77.29,126.47) .. controls (75.59,125.29) and (75.98,123.49) .. (78.15,122.44) -- cycle ;
\draw  [color={rgb, 255:red, 0; green, 0; blue, 0 }  ,draw opacity=1 ][fill={rgb, 255:red, 0; green, 0; blue, 0 }  ,fill opacity=1 ] (37.11,110.77) .. controls (39.29,109.73) and (42.42,109.84) .. (44.12,111.02) .. controls (45.82,112.2) and (45.43,114.01) .. (43.26,115.05) .. controls (41.09,116.09) and (37.95,115.98) .. (36.25,114.8) .. controls (34.55,113.62) and (34.94,111.82) .. (37.11,110.77) -- cycle ;
\draw  [color={rgb, 255:red, 155; green, 155; blue, 155 }  ,draw opacity=0.48 ][fill={rgb, 255:red, 155; green, 155; blue, 155 }  ,fill opacity=0.48 ] (59.07,106.74) .. controls (61.24,105.69) and (64.38,105.8) .. (66.07,106.98) .. controls (67.77,108.16) and (67.39,109.97) .. (65.21,111.01) .. controls (63.04,112.06) and (59.9,111.95) .. (58.2,110.76) .. controls (56.51,109.58) and (56.89,107.78) .. (59.07,106.74) -- cycle ;
\draw  [dash pattern={on 4.5pt off 4.5pt}] (49.18,51.87) .. controls (63.73,44.88) and (84.75,45.62) .. (96.12,53.53) .. controls (107.49,61.44) and (104.91,73.52) .. (90.35,80.51) .. controls (75.8,87.5) and (54.78,86.76) .. (43.41,78.85) .. controls (32.04,70.94) and (34.63,58.86) .. (49.18,51.87) -- cycle ;
\draw  [dash pattern={on 4.5pt off 4.5pt}] (41.55,94.56) .. controls (56.11,87.56) and (77.12,88.31) .. (88.49,96.21) .. controls (99.86,104.12) and (97.28,116.2) .. (82.73,123.19) .. controls (68.17,130.18) and (47.16,129.44) .. (35.79,121.53) .. controls (24.42,113.63) and (27,101.55) .. (41.55,94.56) -- cycle ;
\node[anchor_point] (out1) at (236,110) {};
\draw  [color={rgb, 255:red, 155; green, 155; blue, 155 }  ,draw opacity=0.48 ][fill={rgb, 255:red, 155; green, 155; blue, 155 }  ,fill opacity=0.48 ] (232.42,107.74) .. controls (234.59,106.69) and (237.73,106.8) .. (239.43,107.98) .. controls (241.13,109.16) and (240.74,110.97) .. (238.57,112.01) .. controls (236.39,113.06) and (233.26,112.95) .. (231.56,111.76) .. controls (229.86,110.58) and (230.25,108.78) .. (232.42,107.74) -- cycle ;
\node[anchor_point] (out2) at (243,66) {};
\draw  [color={rgb, 255:red, 155; green, 155; blue, 155 }  ,draw opacity=0.48 ][fill={rgb, 255:red, 155; green, 155; blue, 155 }  ,fill opacity=0.48 ] (240.05,64.05) .. controls (242.22,63.01) and (245.36,63.12) .. (247.06,64.3) .. controls (248.75,65.48) and (248.37,67.28) .. (246.19,68.33) .. controls (244.02,69.37) and (240.88,69.26) .. (239.19,68.08) .. controls (237.49,66.9) and (237.87,65.1) .. (240.05,64.05) -- cycle ;
\draw   (3,40) -- (103,40) -- (117,134) -- (17,134) -- cycle ;
\draw   (176,40) -- (276,40) -- (290,134) -- (190,134) -- cycle ;
\node[block_large](mpn1) at (147.5,110)  {MPN};
\node[block_large](mpn2) at (147.5,60)  {MPN};
\node [text centered] at (147.5, 86.5) (shared) {shared};
\node [text centered] at (60,145) (dims_in) {[n, 3+c]};
\node [text centered] at (233,145) (dims_out) {[$n_{\text{fps}}$, 3+c']};
\node[anchor_point] (p1) at (60,128.5) {};
\draw[line] (p1) to (mpn1);
\node[anchor_point] (p2) at (63.31,87.21) {};
\draw[line] (p2) to (mpn1);
\node[anchor_point] (p3) at (69.17,86.44) {};
\draw[line] (p3) to (mpn2);
\node[anchor_point] (p4) at (71.74,47.01) {};
\draw[line] (p4) to (mpn2);
\draw[line] (shared) to (mpn1);
\draw[line] (shared) to (mpn2);
\draw[arrow] (mpn1) to (out1);
\draw[arrow] (mpn2) to (out2);
\end{tikzpicture}

%% file: images/fe_layer.tex
\tikzstyle{block_large}=[draw, rounded corners=0.1cm, minimum width=0.5cm]
\tikzstyle{arrow} = [->, color=black, >=latex]
\tikzstyle{line} = [-, color=black, >=latex]
\tikzstyle{line_dashed} = [draw, densely dotted, color=black, >=latex]
\tikzstyle{anchor_point}=[text centered]
\tikzstyle{black_point}=[draw, circle, fill=black, inner sep=0pt, minimum size=0.2cm]
\tikzstyle{grey_point}=[draw, circle, fill=gray!50, inner sep=0pt, minimum size=0.2cm]

\begin{tikzpicture}[scale=\tkzscl]
\draw  [color={rgb, 255:red, 0; green, 0; blue, 200 }  ,draw opacity=0.75 ][fill={rgb, 255:red, 0; green, 0; blue, 200 }  ,fill opacity=0.75 ] (62,49.79) .. controls (64.18,48.74) and (67.32,48.86) .. (69.01,50.04) .. controls (70.71,51.22) and (70.32,53.02) .. (68.15,54.06) .. controls (65.98,55.11) and (62.84,55) .. (61.14,53.82) .. controls (59.45,52.64) and (59.83,50.83) .. (62,49.79) -- cycle ;
\draw  [color={rgb, 255:red, 0; green, 0; blue, 200 }  ,draw opacity=0.75 ][fill={rgb, 255:red, 0; green, 0; blue, 200 }  ,fill opacity=0.75 ] (26.68,62.99) .. controls (28.86,61.95) and (32,62.06) .. (33.69,63.24) .. controls (35.39,64.42) and (35,66.22) .. (32.83,67.27) .. controls (30.66,68.31) and (27.52,68.2) .. (25.82,67.02) .. controls (24.13,65.84) and (24.51,64.04) .. (26.68,62.99) -- cycle ;
\draw  [color={rgb, 255:red, 0; green, 0; blue, 200 }  ,draw opacity=0.75 ][fill={rgb, 255:red, 0; green, 0; blue, 200 }  ,fill opacity=0.75 ] (63.22,69.92) .. controls (65.39,68.88) and (68.53,68.99) .. (70.23,70.17) .. controls (71.92,71.35) and (71.54,73.15) .. (69.37,74.2) .. controls (67.19,75.24) and (64.06,75.13) .. (62.36,73.95) .. controls (60.66,72.77) and (61.05,70.97) .. (63.22,69.92) -- cycle ;
\draw  [color={rgb, 255:red, 155; green, 155; blue, 155 }  ,draw opacity=0.48 ][fill={rgb, 255:red, 155; green, 155; blue, 155 }  ,fill opacity=0.48 ] (46.69,64.05) .. controls (48.87,63.01) and (52,63.12) .. (53.7,64.3) .. controls (55.4,65.48) and (55.01,67.28) .. (52.84,68.33) .. controls (50.67,69.37) and (47.53,69.26) .. (45.83,68.08) .. controls (44.13,66.9) and (44.52,65.1) .. (46.69,64.05) -- cycle ;
\draw  [color={rgb, 255:red, 0; green, 0; blue, 200 }  ,draw opacity=0.75 ][fill={rgb, 255:red, 0; green, 0; blue, 200 }  ,fill opacity=0.75 ] (18.42,93.2) .. controls (20.59,92.15) and (23.73,92.26) .. (25.43,93.44) .. controls (27.12,94.62) and (26.74,96.43) .. (24.57,97.47) .. controls (22.39,98.51) and (19.25,98.4) .. (17.56,97.22) .. controls (15.86,96.04) and (16.24,94.24) .. (18.42,93.2) -- cycle ;
\draw  [color={rgb, 255:red, 0; green, 0; blue, 200 }  ,draw opacity=0.75 ][fill={rgb, 255:red, 0; green, 0; blue, 200 }  ,fill opacity=0.75 ] (67.98,61.98) .. controls (70.15,60.94) and (73.29,61.05) .. (74.99,62.23) .. controls (76.69,63.41) and (76.3,65.22) .. (74.13,66.26) .. controls (71.96,67.3) and (68.82,67.19) .. (67.12,66.01) .. controls (65.42,64.83) and (65.81,63.03) .. (67.98,61.98) -- cycle ;
\draw  [color={rgb, 255:red, 0; green, 0; blue, 200 }  ,draw opacity=0.75 ][fill={rgb, 255:red, 0; green, 0; blue, 200 }  ,fill opacity=0.75 ] (37.17,79.93) .. controls (39.34,78.89) and (42.48,79) .. (44.18,80.18) .. controls (45.87,81.36) and (45.49,83.16) .. (43.31,84.21) .. controls (41.14,85.25) and (38,85.14) .. (36.31,83.96) .. controls (34.61,82.78) and (34.99,80.97) .. (37.17,79.93) -- cycle ;
\draw  [color={rgb, 255:red, 0; green, 0; blue, 200 }  ,draw opacity=0.75 ][fill={rgb, 255:red, 0; green, 0; blue, 200 }  ,fill opacity=0.75 ] (61.48,91.97) .. controls (63.65,90.92) and (66.79,91.03) .. (68.49,92.21) .. controls (70.19,93.39) and (69.8,95.2) .. (67.63,96.24) .. controls (65.45,97.29) and (62.32,97.17) .. (60.62,95.99) .. controls (58.92,94.81) and (59.31,93.01) .. (61.48,91.97) -- cycle ;
\draw  [color={rgb, 255:red, 0; green, 0; blue, 200 }  ,draw opacity=0.75 ][fill={rgb, 255:red, 0; green, 0; blue, 200 }  ,fill opacity=0.75 ] (58.15,122.44) .. controls (60.33,121.4) and (63.46,121.51) .. (65.16,122.69) .. controls (66.86,123.87) and (66.47,125.68) .. (64.3,126.72) .. controls (62.13,127.76) and (58.99,127.65) .. (57.29,126.47) .. controls (55.59,125.29) and (55.98,123.49) .. (58.15,122.44) -- cycle ;
\draw  [color={rgb, 255:red, 0; green, 0; blue, 200 }  ,draw opacity=0.75 ][fill={rgb, 255:red, 0; green, 0; blue, 200 }  ,fill opacity=0.75 ] (17.11,110.77) .. controls (19.29,109.73) and (22.42,109.84) .. (24.12,111.02) .. controls (25.82,112.2) and (25.43,114.01) .. (23.26,115.05) .. controls (21.09,116.09) and (17.95,115.98) .. (16.25,114.8) .. controls (14.55,113.62) and (14.94,111.82) .. (17.11,110.77) -- cycle ;
\draw  [color={rgb, 255:red, 155; green, 155; blue, 155 }  ,draw opacity=0.48 ][fill={rgb, 255:red, 155; green, 155; blue, 155 }  ,fill opacity=0.48 ] (39.07,106.74) .. controls (41.24,105.69) and (44.38,105.8) .. (46.07,106.98) .. controls (47.77,108.16) and (47.39,109.97) .. (45.21,111.01) .. controls (43.04,112.06) and (39.9,111.95) .. (38.2,110.76) .. controls (36.51,109.58) and (36.89,107.78) .. (39.07,106.74) -- cycle ;
\draw  [color={rgb, 255:red, 0; green, 0; blue, 0 }  ,draw opacity=1 ][fill={rgb, 255:red, 0; green, 0; blue, 0 }  ,fill opacity=1 ] (92,-45.21) .. controls (94.18,-46.26) and (97.32,-46.14) .. (99.01,-44.96) .. controls (100.71,-43.78) and (100.32,-41.98) .. (98.15,-40.94) .. controls (95.98,-39.89) and (92.84,-40) .. (91.14,-41.18) .. controls (89.45,-42.36) and (89.83,-44.17) .. (92,-45.21) -- cycle ;
\draw  [color={rgb, 255:red, 0; green, 0; blue, 0 }  ,draw opacity=1 ][fill={rgb, 255:red, 0; green, 0; blue, 0 }  ,fill opacity=1 ] (56.68,-32.01) .. controls (58.86,-33.05) and (62,-32.94) .. (63.69,-31.76) .. controls (65.39,-30.58) and (65,-28.78) .. (62.83,-27.73) .. controls (60.66,-26.69) and (57.52,-26.8) .. (55.82,-27.98) .. controls (54.13,-29.16) and (54.51,-30.96) .. (56.68,-32.01) -- cycle ;
\draw  [color={rgb, 255:red, 0; green, 0; blue, 0 }  ,draw opacity=1 ][fill={rgb, 255:red, 0; green, 0; blue, 0 }  ,fill opacity=1 ] (93.22,-25.08) .. controls (95.39,-26.12) and (98.53,-26.01) .. (100.23,-24.83) .. controls (101.92,-23.65) and (101.54,-21.85) .. (99.37,-20.8) .. controls (97.19,-19.76) and (94.06,-19.87) .. (92.36,-21.05) .. controls (90.66,-22.23) and (91.05,-24.03) .. (93.22,-25.08) -- cycle ;
\draw  [color={rgb, 255:red, 0; green, 0; blue, 0 }  ,draw opacity=1 ][fill={rgb, 255:red, 0; green, 0; blue, 0 }  ,fill opacity=1 ] (76.69,-30.95) .. controls (78.87,-31.99) and (82,-31.88) .. (83.7,-30.7) .. controls (85.4,-29.52) and (85.01,-27.72) .. (82.84,-26.67) .. controls (80.67,-25.63) and (77.53,-25.74) .. (75.83,-26.92) .. controls (74.13,-28.1) and (74.52,-29.9) .. (76.69,-30.95) -- cycle ;
\draw  [color={rgb, 255:red, 0; green, 0; blue, 0 }  ,draw opacity=1 ][fill={rgb, 255:red, 0; green, 0; blue, 0 }  ,fill opacity=1 ] (48.42,-1.8) .. controls (50.59,-2.85) and (53.73,-2.74) .. (55.43,-1.56) .. controls (57.12,-0.38) and (56.74,1.43) .. (54.57,2.47) .. controls (52.39,3.51) and (49.25,3.4) .. (47.56,2.22) .. controls (45.86,1.04) and (46.24,-0.76) .. (48.42,-1.8) -- cycle ;
\draw  [color={rgb, 255:red, 0; green, 0; blue, 0 }  ,draw opacity=1 ][fill={rgb, 255:red, 0; green, 0; blue, 0 }  ,fill opacity=1 ] (97.98,-33.02) .. controls (100.15,-34.06) and (103.29,-33.95) .. (104.99,-32.77) .. controls (106.69,-31.59) and (106.3,-29.78) .. (104.13,-28.74) .. controls (101.96,-27.7) and (98.82,-27.81) .. (97.12,-28.99) .. controls (95.42,-30.17) and (95.81,-31.97) .. (97.98,-33.02) -- cycle ;
\draw  [color={rgb, 255:red, 0; green, 0; blue, 0 }  ,draw opacity=1 ][fill={rgb, 255:red, 0; green, 0; blue, 0 }  ,fill opacity=1 ] (67.17,-15.07) .. controls (69.34,-16.11) and (72.48,-16) .. (74.18,-14.82) .. controls (75.87,-13.64) and (75.49,-11.84) .. (73.31,-10.79) .. controls (71.14,-9.75) and (68,-9.86) .. (66.31,-11.04) .. controls (64.61,-12.22) and (64.99,-14.03) .. (67.17,-15.07) -- cycle ;
\draw  [color={rgb, 255:red, 0; green, 0; blue, 0 }  ,draw opacity=1 ][fill={rgb, 255:red, 0; green, 0; blue, 0 }  ,fill opacity=1 ] (91.48,-3.03) .. controls (93.65,-4.08) and (96.79,-3.97) .. (98.49,-2.79) .. controls (100.19,-1.61) and (99.8,0.2) .. (97.63,1.24) .. controls (95.45,2.29) and (92.32,2.17) .. (90.62,0.99) .. controls (88.92,-0.19) and (89.31,-1.99) .. (91.48,-3.03) -- cycle ;
\draw  [color={rgb, 255:red, 0; green, 0; blue, 0 }  ,draw opacity=1 ][fill={rgb, 255:red, 0; green, 0; blue, 0 }  ,fill opacity=1 ] (88.15,27.44) .. controls (90.33,26.4) and (93.46,26.51) .. (95.16,27.69) .. controls (96.86,28.87) and (96.47,30.68) .. (94.3,31.72) .. controls (92.13,32.76) and (88.99,32.65) .. (87.29,31.47) .. controls (85.59,30.29) and (85.98,28.49) .. (88.15,27.44) -- cycle ;
\draw  [color={rgb, 255:red, 0; green, 0; blue, 0 }  ,draw opacity=1 ][fill={rgb, 255:red, 0; green, 0; blue, 0 }  ,fill opacity=1 ] (47.11,15.77) .. controls (49.29,14.73) and (52.42,14.84) .. (54.12,16.02) .. controls (55.82,17.2) and (55.43,19.01) .. (53.26,20.05) .. controls (51.09,21.09) and (47.95,20.98) .. (46.25,19.8) .. controls (44.55,18.62) and (44.94,16.82) .. (47.11,15.77) -- cycle ;
\draw  [color={rgb, 255:red, 0; green, 0; blue, 0 }  ,draw opacity=1 ][fill={rgb, 255:red, 0; green, 0; blue, 0 }  ,fill opacity=1 ] (69.07,11.74) .. controls (71.24,10.69) and (74.38,10.8) .. (76.07,11.98) .. controls (77.77,13.16) and (77.39,14.97) .. (75.21,16.01) .. controls (73.04,17.06) and (69.9,16.95) .. (68.2,15.76) .. controls (66.51,14.58) and (66.89,12.78) .. (69.07,11.74) -- cycle ;
\draw  [color={rgb, 255:red, 97; green, 167; blue, 220}  ,draw opacity=1 ][fill={rgb, 255:red, 97; green, 167; blue, 220}  ,fill opacity=1 ] (62+200,49.79-47.5) .. controls (64.18+200,48.74-47.5) and (67.32+200,48.86-47.5) .. (69.01+200,50.04-47.5) .. controls (70.71+200,51.22-47.5) and (70.32+200,53.02-47.5) .. (68.15+200,54.06-47.5) .. controls (65.98+200,55.11-47.5) and (62.84+200,55-47.5) .. (61.14+200,53.82-47.5) .. controls (59.45+200,52.64-47.5) and (59.83+200,50.83-47.5) .. (62+200,49.79-47.5) -- cycle ;
\draw  [color={rgb, 255:red, 97; green, 167; blue, 220}  ,draw opacity=1 ][fill={rgb, 255:red, 97; green, 167; blue, 220}  ,fill opacity=1 ] (26.68+200,62.99-47.5) .. controls (28.86+200,61.95-47.5) and (32+200,62.06-47.5) .. (33.69+200,63.24-47.5) .. controls (35.39+200,64.42-47.5) and (35+200,66.22-47.5) .. (32.83+200,67.27-47.5) .. controls (30.66+200,68.31-47.5) and (27.52+200,68.2-47.5) .. (25.82+200,67.02-47.5) .. controls (24.13+200,65.84-47.5) and (24.51+200,64.04-47.5) .. (26.68+200,62.99-47.5) -- cycle ;
\draw  [color={rgb, 255:red, 97; green, 167; blue, 220}  ,draw opacity=1 ][fill={rgb, 255:red, 97; green, 167; blue, 220}  ,fill opacity=1 ] (63.22+200,69.92-47.5) .. controls (65.39+200,68.88-47.5) and (68.53+200,68.99-47.5) .. (70.23+200,70.17-47.5) .. controls (71.92+200,71.35-47.5) and (71.54+200,73.15-47.5) .. (69.37+200,74.2-47.5) .. controls (67.19+200,75.24-47.5) and (64.06+200,75.13-47.5) .. (62.36+200,73.95-47.5) .. controls (60.66+200,72.77-47.5) and (61.05+200,70.97-47.5) .. (63.22+200,69.92-47.5) -- cycle ;
\draw  [color={rgb, 255:red, 155; green, 155; blue, 155 }  ,draw opacity=0.48 ][fill={rgb, 255:red, 155; green, 155; blue, 155 }  ,fill opacity=0.48 ] (46.69+200,64.05-47.5) .. controls (48.87+200,63.01-47.5) and (52+200,63.12-47.5) .. (53.7+200,64.3-47.5) .. controls (55.4+200,65.48-47.5) and (55.01+200,67.28-47.5) .. (52.84+200,68.33-47.5) .. controls (50.67+200,69.37-47.5) and (47.53+200,69.26-47.5) .. (45.83+200,68.08-47.5) .. controls (44.13+200,66.9-47.5) and (44.52+200,65.1-47.5) .. (46.69+200,64.05-47.5) -- cycle ;
\draw  [color={rgb, 255:red, 97; green, 167; blue, 220}  ,draw opacity=1 ][fill={rgb, 255:red, 97; green, 167; blue, 220}  ,fill opacity=1 ] (18.42+200,93.2-47.5) .. controls (20.59+200,92.15-47.5) and (23.73+200,92.26-47.5) .. (25.43+200,93.44-47.5) .. controls (27.12+200,94.62-47.5) and (26.74+200,96.43-47.5) .. (24.57+200,97.47-47.5) .. controls (22.39+200,98.51-47.5) and (19.25+200,98.4-47.5) .. (17.56+200,97.22-47.5) .. controls (15.86+200,96.04-47.5) and (16.24+200,94.24-47.5) .. (18.42+200,93.2-47.5) -- cycle ;
\draw  [color={rgb, 255:red, 97; green, 167; blue, 220}  ,draw opacity=1 ][fill={rgb, 255:red, 97; green, 167; blue, 220}  ,fill opacity=1 ] (67.98+200,61.98-47.5) .. controls (70.15+200,60.94-47.5) and (73.29+200,61.05-47.5) .. (74.99+200,62.23-47.5) .. controls (76.69+200,63.41-47.5) and (76.3+200,65.22-47.5) .. (74.13+200,66.26-47.5) .. controls (71.96+200,67.3-47.5) and (68.82+200,67.19-47.5) .. (67.12+200,66.01-47.5) .. controls (65.42+200,64.83-47.5) and (65.81+200,63.03-47.5) .. (67.98+200,61.98-47.5) -- cycle ;
\draw  [color={rgb, 255:red, 97; green, 167; blue, 220}  ,draw opacity=1 ][fill={rgb, 255:red, 97; green, 167; blue, 220}  ,fill opacity=1 ] (37.17+200,79.93-47.5) .. controls (39.34+200,78.89-47.5) and (42.48+200,79-47.5) .. (44.18+200,80.18-47.5) .. controls (45.87+200,81.36-47.5) and (45.49+200,83.16-47.5) .. (43.31+200,84.21-47.5) .. controls (41.14+200,85.25-47.5) and (38+200,85.14-47.5) .. (36.31+200,83.96-47.5) .. controls (34.61+200,82.78-47.5) and (34.99+200,80.97-47.5) .. (37.17+200,79.93-47.5) -- cycle ;
\draw  [color={rgb, 255:red, 97; green, 167; blue, 220}  ,draw opacity=1 ][fill={rgb, 255:red, 97; green, 167; blue, 220}  ,fill opacity=1 ] (61.48+200,91.97-47.5) .. controls (63.65+200,90.92-47.5) and (66.79+200,91.03-47.5) .. (68.49+200,92.21-47.5) .. controls (70.19+200,93.39-47.5) and (69.8+200,95.2-47.5) .. (67.63+200,96.24-47.5) .. controls (65.45+200,97.29-47.5) and (62.32+200,97.17-47.5) .. (60.62+200,95.99-47.5) .. controls (58.92+200,94.81-47.5) and (59.31+200,93.01-47.5) .. (61.48+200,91.97-47.5) -- cycle ;
\draw  [color={rgb, 255:red, 97; green, 167; blue, 220}  ,draw opacity=1 ][fill={rgb, 255:red, 97; green, 167; blue, 220}  ,fill opacity=1 ] (58.15+200,122.44-47.5) .. controls (60.33+200,121.4-47.5) and (63.46+200,121.51-47.5) .. (65.16+200,122.69-47.5) .. controls (66.86+200,123.87-47.5) and (66.47+200,125.68-47.5) .. (64.3+200,126.72-47.5) .. controls (62.13+200,127.76-47.5) and (58.99+200,127.65-47.5) .. (57.29+200,126.47-47.5) .. controls (55.59+200,125.29-47.5) and (55.98+200,123.49-47.5) .. (58.15+200,122.44-47.5) -- cycle ;
\draw  [color={rgb, 255:red, 97; green, 167; blue, 220}  ,draw opacity=1 ][fill={rgb, 255:red, 97; green, 167; blue, 220}  ,fill opacity=1 ] (17.11+200,110.77-47.5) .. controls (19.29+200,109.73-47.5) and (22.42+200,109.84-47.5) .. (24.12+200,111.02-47.5) .. controls (25.82+200,112.2-47.5) and (25.43+200,114.01-47.5) .. (23.26+200,115.05-47.5) .. controls (21.09+200,116.09-47.5) and (17.95+200,115.98-47.5) .. (16.25+200,114.8-47.5) .. controls (14.55+200,113.62-47.5) and (14.94+200,111.82-47.5) .. (17.11+200,110.77-47.5) -- cycle ;
\draw  [color={rgb, 255:red, 155; green, 155; blue, 155 }  ,draw opacity=0.48 ][fill={rgb, 255:red, 155; green, 155; blue, 155 }  ,fill opacity=0.48 ] (39.07+200,106.74-47.5) .. controls (41.24+200,105.69-47.5) and (44.38+200,105.8-47.5) .. (46.07+200,106.98-47.5) .. controls (47.77+200,108.16-47.5) and (47.39+200,109.97-47.5) .. (45.21+200,111.01-47.5) .. controls (43.04+200,112.06-47.5) and (39.9+200,111.95-47.5) .. (38.2+200,110.76-47.5) .. controls (36.51+200,109.58-47.5) and (36.89+200,107.78-47.5) .. (39.07+200,106.74-47.5) -- cycle ;
\draw  [dash pattern={on 4.5pt off 4.5pt}] (49.18-20,51.87-95) .. controls (63.73-20,44.88-95) and (84.75-20,45.62-95) .. (96.12-20,53.53-95) .. controls (107.49-20,61.44-95) and (104.91-20,73.52-95) .. (90.35-20,80.51-95) .. controls (75.8-20,87.5-95) and (54.78-20,86.76-95) .. (43.41-20,78.85-95) .. controls (32.04-20,70.94-95) and (34.63-20,58.86-95) .. (49.18-20,51.87-95) -- cycle ;
\draw  [dash pattern={on 4.5pt off 4.5pt}] (41.55-20,94.56-95) .. controls (56.11-20,87.56-95) and (77.12-20,88.31-95) .. (88.49-20,96.21-95) .. controls (99.86-20,104.12-95) and (97.28-20,116.2-95) .. (82.73-20,123.19-95) .. controls (68.17-20,130.18-95) and (47.16-20,129.44-95) .. (35.79-20,121.53-95) .. controls (24.42-20,113.63-95) and (27-20,101.55-95) .. (41.55-20,94.56-95) -- cycle ;
\draw   (3-10,45) -- (103+5,45) -- (117+5,130) -- (17-10,130) -- cycle ;
\draw   (3-10,-50) -- (103+5,-50) -- (117+5,35) -- (17-10,35) -- cycle ;
\draw   (3-10+200,45-47.5) -- (103+5+200,45-47.5) -- (117+5+200,130-47.5) -- (17-10+200,130-47.5) -- cycle ;
\node[block_large](mpn1) at (147.5,110-95)  {MPN};
\node[block_large](mpn2) at (147.5,60-95)  {MPN};
\node [text centered] at (147.5, 86.5-95) (shared) {shared};
\node [text centered] at (60,145) (dims_in) {[$n_1$, 3+c]};
\node [text centered] at (60,145-115-95) (dims_in) {[$n_2$, 3+c]};
\node [text centered] at (60+200,145-47.5) (dims_out) {[$n_1$, 3+c']};
\node[anchor_point] (p1) at (42,113) {};
\node[anchor_point] (p2) at (42,110-98) {};
\draw[line_dashed] (p1) to (p2);
\node[anchor_point] (p3) at (46+4,64+7) {};
\node[anchor_point] (p4) at (46+4,64-7-95) {};
\draw[line_dashed] (p3) to (p4);
\node[anchor_point] (p5) at (42-5,110-98+22) {};
\node[anchor_point] (p6) at (42-5,110-98-18) {};
\draw[line] (p5) to (mpn1);
\draw[line] (p6) to (mpn1);
\node[anchor_point] (p7) at (46,64-95+22) {};
\node[anchor_point] (p8) at (46,64-95-18) {};
\draw[line] (p7) to (mpn2);
\draw[line] (p8) to (mpn2);
\draw[line] (shared) to (mpn1);
\draw[line] (shared) to (mpn2);
\node[anchor_point] (out1) at (43+200,106.74-45) {};
\draw[arrow] (mpn1) to (out1);
\node[anchor_point] (out2) at (50+200,64.05-45) {};
\draw[arrow] (mpn2) to (out2);
\node [text centered] at (60+45,145-25) (label_in_P) {$P$};
\node [text centered] at (60-55,145-115-95+25) (label_in_Q) {$Q$};
\node [text centered] at (60+200+45,145-47.5-25) (label_out_P) {$P'$};
\end{tikzpicture}